\documentclass[11pt,a4paper]{article}
\usepackage{times,latexsym}
\usepackage{url}
\usepackage[T1]{fontenc}

\usepackage[acceptedWithA]{tacl2021v1}

\usepackage{xspace,mfirstuc,tabulary}

\usepackage{graphicx}
\usepackage{amsmath}
\usepackage{blindtext}
\usepackage{booktabs}
\usepackage{multirow}
\usepackage{cleveref}
\usepackage{xr}
\usepackage{subcaption}
\usepackage{tablefootnote}

\newcommand{\repr}[1]{\textbf{#1}}
\makeatletter
\newcommand*{\rom}[1]{\expandafter\@slowromancap\romannumeral #1@}
\makeatother

\newif\iftaclinstructions
\taclinstructionsfalse 
\iftaclinstructions
\renewcommand{\confidential}{}
\renewcommand{\anonsubtext}{(No author info supplied here, for consistency with
TACL-submission anonymization requirements)}
\newcommand{\instr}
\fi

\iftaclpubformat 

\else

\fi

\title{Building an Efficiency Pipeline: Commutativity and Cumulativeness of Efficiency Operators for Transformers}

\author{Ji Xin, Raphael Tang, Zhiying Jiang, Yaoliang Yu, \and Jimmy Lin \vspace{5pt} \\
	David R. Cheriton School of Computer Science \\ University of Waterloo \\
	{\tt   \{ji.xin,r33tang,zhiying.jiang,yaoliang.yu,jimmylin\}@uwaterloo.ca}
}

\date{}

\begin{document}
\maketitle

\begin{abstract}
There exists a wide variety of efficiency methods for natural language processing (NLP) tasks, such as pruning, distillation, dynamic inference, quantization, etc.
We can consider an efficiency method as an \textit{operator} applied on a model.
Naturally, we may construct a pipeline of multiple efficiency methods, i.e., to apply multiple operators on the model sequentially.
In this paper, we study the plausibility of this idea, and more importantly, the \textit{commutativity} and \textit{cumulativeness} of efficiency operators.
We make two interesting observations:
(1) Efficiency operators are commutative---the order of efficiency methods within the pipeline has little impact on the final results;
(2) Efficiency operators are also cumulative---the final results of combining several efficiency methods can be estimated by combining the results of individual methods.
These observations deepen our understanding of efficiency operators and provide useful guidelines for their real-world applications.
\end{abstract}

\section{Introduction}
\label{sec:intro}

Natural language processing (NLP) tasks nowadays heavily rely on complex neural models, especially large-scale pre-trained language models based on the transformer architecture~\cite{transformer}, such as BERT~\cite{bert} and RoBERTa~\cite{roberta}.
Despite being more accurate than previous models, transformer-based models are typically slow to execute, making it a non-trivial challenge to apply them in real-world applications.
For example, it takes a BERT-base model about 200 ms per query to perform a simple sequence classification task on a commercial CPU, which can be too slow in many scenarios.
Therefore, model efficiency has become an increasingly important research direction in the transformer era.

A wide variety of efficiency methods have been individually studied for transformers, like pruning~\cite{pruning_bert}, distillation~\cite{distilbert}, dynamic inference~\cite{deebert,lat}, and quantization~\cite{qbert}, to name a few.
There has also been work on applying multiple efficiency methods together as a pipeline~\cite{fastformers,lin-etal-2021-bag-tricks,CUI202156}, but the construction of such pipelines has not been methodically studied.
It remains unclear how to choose components for the pipeline among numerous options, since time savings often come at the price of accuracy drops, and therefore, na\"ively stacking all available efficiency methods leads to poor performance.
Furthermore, even with a chosen set of efficiency methods, different orders of applying them may yield different results and it is nontrivial to find the best order.

In this paper, we study how to effectively construct a pipeline of efficiency methods.
Conceptually, we consider each efficiency method as an \textit{operator} applied on a model and study the properties of these efficiency operators.
We conduct experiments with the RoBERTa model~\cite{roberta} on a number of NLP tasks and include the following components in our efficiency pipelines:\ distillation, structured pruning, quantization, early exiting, and dynamic length inference.
We study two important properties of efficiency operators: (1) Commutativity: does arbitrarily swapping the order of operators affect the final accuracy--efficiency tradeoff of the model? (2) Cumulativeness: how do the two metrics, time savings and accuracy drops, compound across multiple operators?
For commutativity, we show that the difference between various orderings of the same set of components is usually small and negligible in practice.
For cumulativeness, we show that time saving and accuracy drop are both cumulative to the extent that we can estimate the performance of a new pipeline by combining the results of individual components.
These observations make it more convenient for us to build new pipelines and estimate their performance without having to carry out time-consuming experiments.

\section{Related Work}

In this section, we discuss related work, including individual efficiency methods and combining multiple ones.

\subsection{Individual Efficiency Methods}
Efficiency methods for machine learning have been studied for many decades.
We briefly introduce the most common ones in the context of transformer models.

\paragraph{Knowledge distillation}
Distillation~\cite{distillation} aims to distill knowledge from a large and costly \textit{teacher} model to a small and efficient \textit{student} model.
\citet{distilltolstm} perform task-specific distillation from a fine-tuned BERT model into non-transformer architectures such as LSTMs aligning predicted logits of the teacher and the student.
Patient knowledge distillation~\cite{pkd} performs task-specific distillation, where the students are transformer models with smaller depth and width;
furthermore, they align not only predicted logits but also intermediate states of both models.
DistilBERT~\cite{distilbert} and TinyBERT~\cite{tinybert} perform both task-agnostic and task-specific distillation:\ first the student model learns from a pre-trained teacher;
then it can either be directly fine-tuned like a pre-trained model or learn from another fine-tuned teacher as a student.

\paragraph{Pruning}

Structured pruning~\cite{pruning,structuredpruning,compressingbert} removes high-level components of a model, such as an attention head or an entire row/column in a feed-forward network (FFN)'s weight matrix, and can be directly used for improving model efficiency.
\citet{sixteenheads} show that reducing attention heads \textit{after} training/fine-tuning does not significantly degrade the model's effectiveness and argue that in a lot of cases, the number of attention heads can be reduced.
MobileBERT~\cite{mobilebert} reduces the intermediate dimension of a transformer layer's FFN by using a funnel-like structure to first shrink the intermediate layer size and then recover it at the end of the layer.
\citet{pruning_bert} improves BERT efficiency for question answering by reducing both attention heads and intermediate dimensions.

\paragraph{Dynamic Depth and Length}
For dynamic depth, early exiting~\cite{deebert,fastbert} converts the original fine-tuned model into a multi-output one, and dynamically chooses the number of layers used for the inference of each example, based on model confidence~\cite{righttool}, model patience~\cite{pabee}, or the prediction of an external controlling module~\cite{berxit}.
For dynamic length, PoWER-BERT~\cite{powerbert} shrinks the sequence length gradually as going into deep layers, eventually reducing the sequence length to $1$ at the final layer for sequence-level prediction;
Length-Adaptive Transformer~\cite{lat} extends the idea to token-level prediction by first reducing the sequence length and then recovering missing tokens' outputs.

\subsection{Applying Multiple Efficiency Methods}
There has been work on both reducing multiple dimensions of a model and applying multiple efficiency methods in a pipeline.
DynaBERT~\cite{dynabert} improves model efficiency by first reducing model width and then reducing depth.
\citet{CUI202156} perform pruning and distillation jointly for model compression.
\citet{lin-etal-2021-bag-tricks} propose a bag of tricks to accelerate the inference stage of neural machine translation models.
Fastformers~\cite{fastformers} propose a pipeline consisting of several components which together provide more than $100\times$ acceleration.
Despite the success of constructing an efficiency pipeline, it remains under\-explored \textit{how} these pipelines should be built in order to achieve the best accuracy--efficiency tradeoffs.
We aim to tackle this problem in our paper.

\begin{figure*}[ht]
    \begin{subfigure}[t]{0.38\textwidth}
    	\includegraphics[height=140pt]{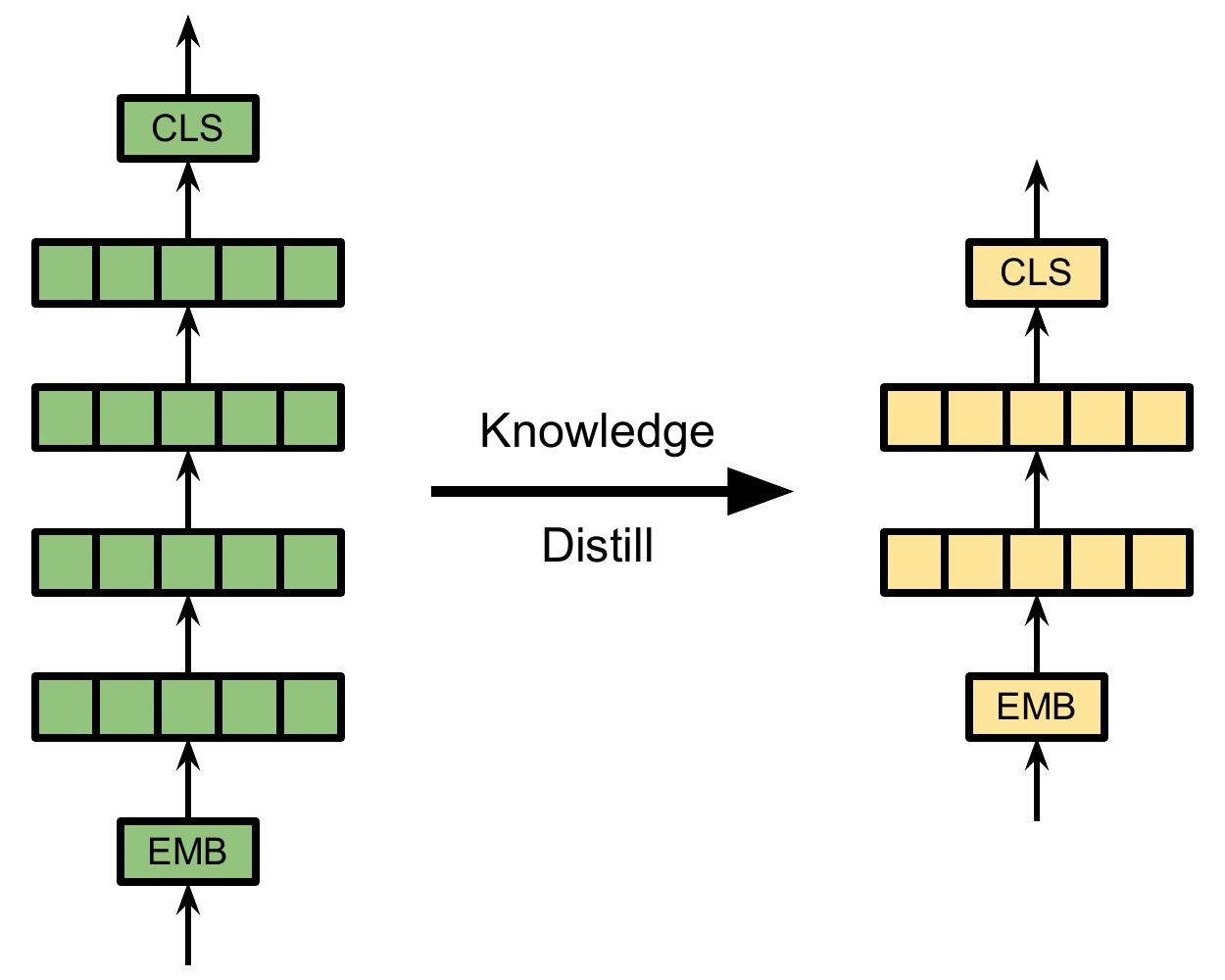}
    	\caption{Distillation}
    	\label{fig:distillation}
    \end{subfigure}
    \begin{subfigure}[t]{0.38\textwidth}
    	\includegraphics[height=140pt]{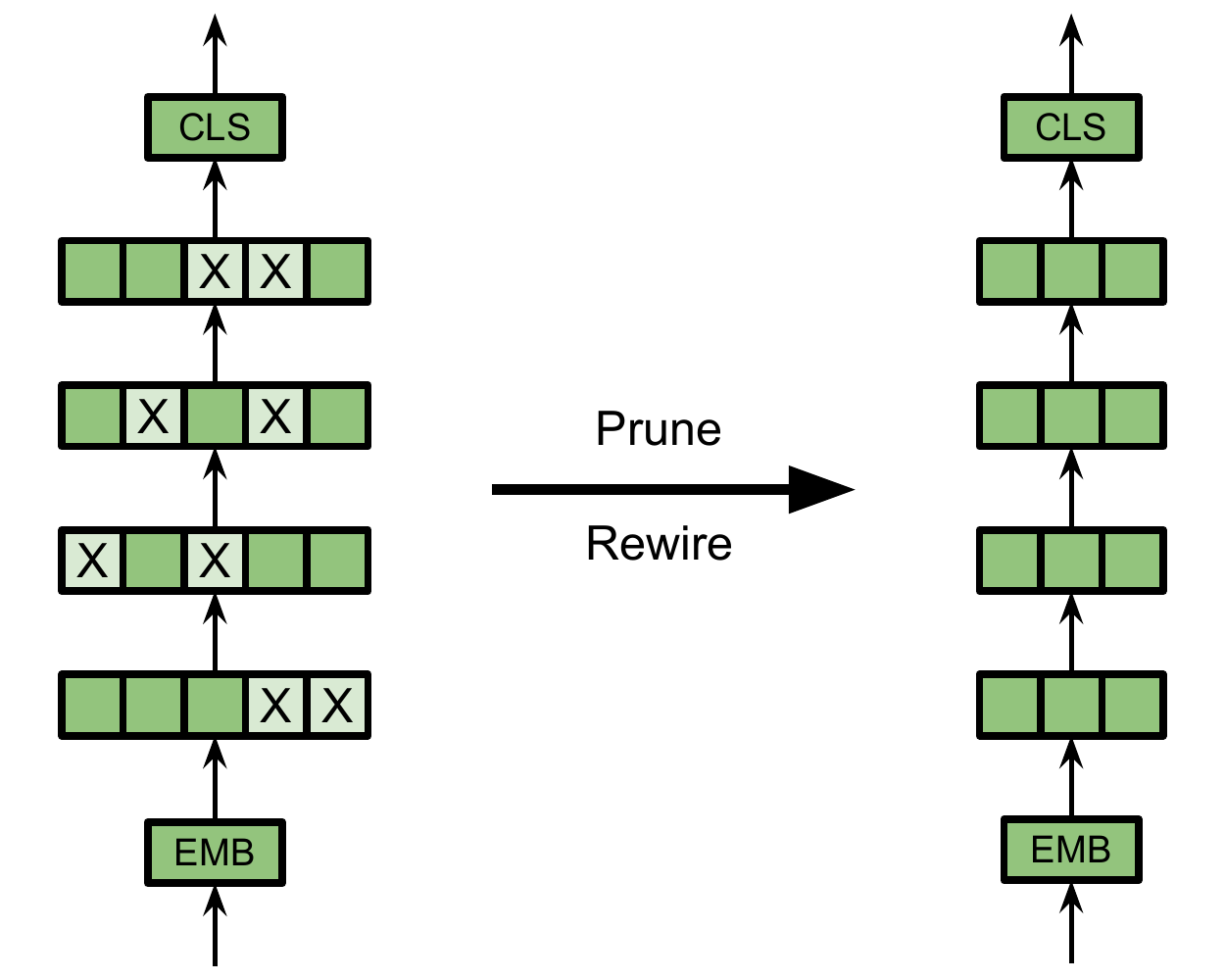}
    	\caption{Pruning}
    	\label{fig:pruning}
    \end{subfigure}
    \begin{subfigure}[t]{0.22\textwidth}
    	\includegraphics[height=140pt]{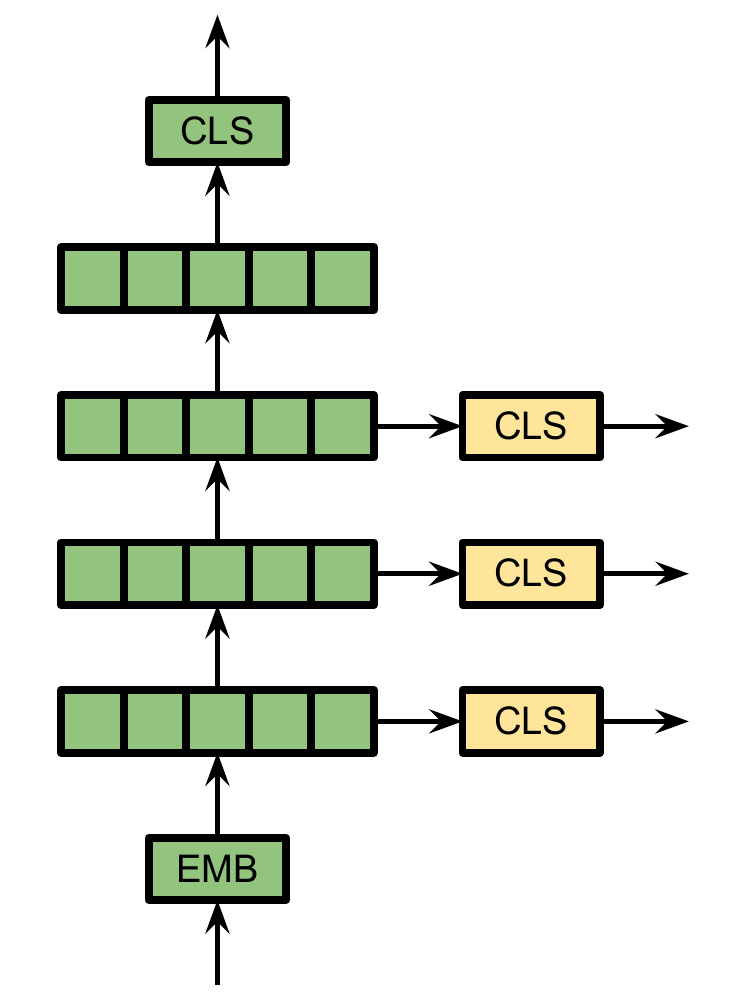}
    	\caption{Early exiting}
    	\label{fig:earlyexiting}
    \end{subfigure}
	\caption{Diagrams for efficiency methods. The model consists of an \textit{embedding layer} (EMB) at the bottom, a few \textit{transformer layers} in the middle, and a \textit{classifier} (CLS) at the top. Green blocks represent parameters available from fine-tuning, and yellow blocks represent parameters that are initialized and optimized after fine-tuning. Distillation initialized a new student model and distill knowledge from the original teacher model; pruning removes unimportant parts of the original model and rewires the connection; early exiting adds extra classifiers for intermediate transformer layers. Dynamic sequence length and quantization are not shown because they do not change the model architecture.}
	\label{fig:model}
\end{figure*}

\section{Efficiency Methods}

In this section, we discuss efficiency methods used in this paper and their modeling choices.
Applying transformers for NLP tasks typically involves three stages:\ pre-training, fine-tuning, and inference~\cite{gpt,bert}.
In this paper, we assume the availability of pre-trained models and study different ways of fine-tuning them to achieve better tradeoffs between inference accuracy and efficiency.
\textit{Training} henceforth refers to fine-tuning in the paper.

\subsection{Distillation}
\label{sec:distill}

Distillation improves efficiency by initializing a smaller model called a \textit{student}, and then distilling knowledge from the original \textit{teacher} model by using the teacher's output as the supervision signal for the student's training.

In the case of transformers, there are two types of distillation, namely task-agnostic and task-specific, depending on whether the student model is trained for a specific task.
These two types correspond to the pre-training stage and the fine-tuning stage.

In this paper, we focus on \textit{task-specific distillation}, which corresponds to fine-tuning (\Cref{fig:distillation}).
We initialize the student model with a TinyBERT~\cite{tinybert} backbone that comes from task-agnostic distillation.
In addition to the most common loss function (teacher supervising student), which is a soft cross-entropy between output logits of the teacher and the student, we introduce two other parts for the loss function: (1) mean squared error (MSE) between the teacher's and the student's embedding layers' outputs; (2) MSE between the teacher's and the student's final transformer layers' outputs.
It has been shown in related work that adding objectives to align intermediate states of the teacher and the student helps with distillation~\cite{pkd,distilbert}.
We simply use a ratio of $1:1:1$ for these three parts of the loss function.

\subsection{Structured Pruning}

Pruning removes unimportant parts of the model and increases the sparsity level of the model.
A specific category of pruning, \textit{structured pruning}, removes high-level units of the model, such as a layer, an attention head, or a group of parameters in the weight matrix.
Model sparsity induced by structured pruning can directly translate to faster execution, and therefore we focus on structured pruning in the paper.

Following the work by \citet{pruning_bert} and \citet{fastformers}, we choose two aspects of the model and prune them separately:\ the number of attention heads and the intermediate dimension of the fully connected layer within a transformer layer (\Cref{fig:pruning}).
We calculate the \textit{importance} of attention heads and intermediate dimensions with a first order method:\ run inference for the entire dev set and accumulate the first order gradients for each attention head and intermediate dimension.
We then remove the least important attention heads and intermediate dimensions, according to the desired sparsity level, and then rewire the model connection so it becomes a smaller but complete model.
After pruning, we perform another round of knowledge distillation from the original model to the pruned model as described in the previous subsection, which further improves the pruned model's accuracy without sacrificing efficiency.

\subsection{Dynamic Inference}

Dynamic inference~\cite{branchynet,adaptivecomputation,ut} accelerates inference by reducing the amount of computation adaptively, depending on the nature of the input example.
We discuss two types of dynamic inference in this section.

\subsubsection{Dynamic Depth: Early Exiting}

Early exiting aims to reduce the number of transformer layers used for inference.
It modifies a fine-tuned model by adding extra classifiers to intermediate transformer layers (\Cref{fig:earlyexiting}).

In order to use these extra classifiers, we further train the model by minimizing the sum of loss functions of all classifiers.
The loss function has the same form for all classifiers:\ the cross entropy between ground truth labels and the classifier's prediction logits.

A special case to notice here is how to perform distillation and pruning after adding early exiting.
\begin{itemize}
    \item Distillation after early exiting. When we initialize the student model (e.g., from TinyBERT), we also add early exiting classifiers to it. For training, the $i^{\text{th}}$ layer of the student model uses the prediction from the $2i^{\text{th}}$ layer of the teacher model as supervision.
    \item Pruning after early exiting. When we prune the transformer layers, we do not change the classifiers. For the additional round of distillation, each layer of the student model uses the prediction from its corresponding layer of the teacher model as supervision.
\end{itemize}

\noindent For inference, the early exiting model sequentially produces an output at each layer's classifier.
If the confidence of a certain layer's output exceeds a preset \textit{early exiting threshold}, the model immediately returns the current layer's output and no longer executes the remaining layers, thereby saving inference computation.

\subsubsection{Dynamic Sequence Length}

Pre-trained language models come with a fixed input sequence length (e.g., 512 for RoBERTa) that aligns with the design of positional embeddings~\cite{bert}.
Inputs longer than the fixed length are truncated and shorter inputs are padded with zero vectors.
This fixed length, while being useful for tasks with long inputs, is often unnecessarily large for most downstream applications.
In this paper, we use a simple method for length reduction:\ for each batch, we dynamically set the input sequence length to the maximum length of inputs within the batch.
This reduces the number of zero paddings in input sequences and reduces unnecessary computation.
Different from previous methods, dynamic sequence length does not affect the model's accuracy.

\subsection{Quantization}

Quantization~\cite{quantization,qbert} improves model efficiency by using fewer bits to store and process data.
The idea itself is straightforward, but implementation can be highly hardware dependent.
Since we run inference on CPUs, we first export the trained model to ONNX\footnote{\url{https://onnx.ai/}.} and then run it with 8-bit quantization, following Fastformers~\cite{fastformers}.

\begin{figure*}[ht]
	\includegraphics[width=0.33\textwidth]{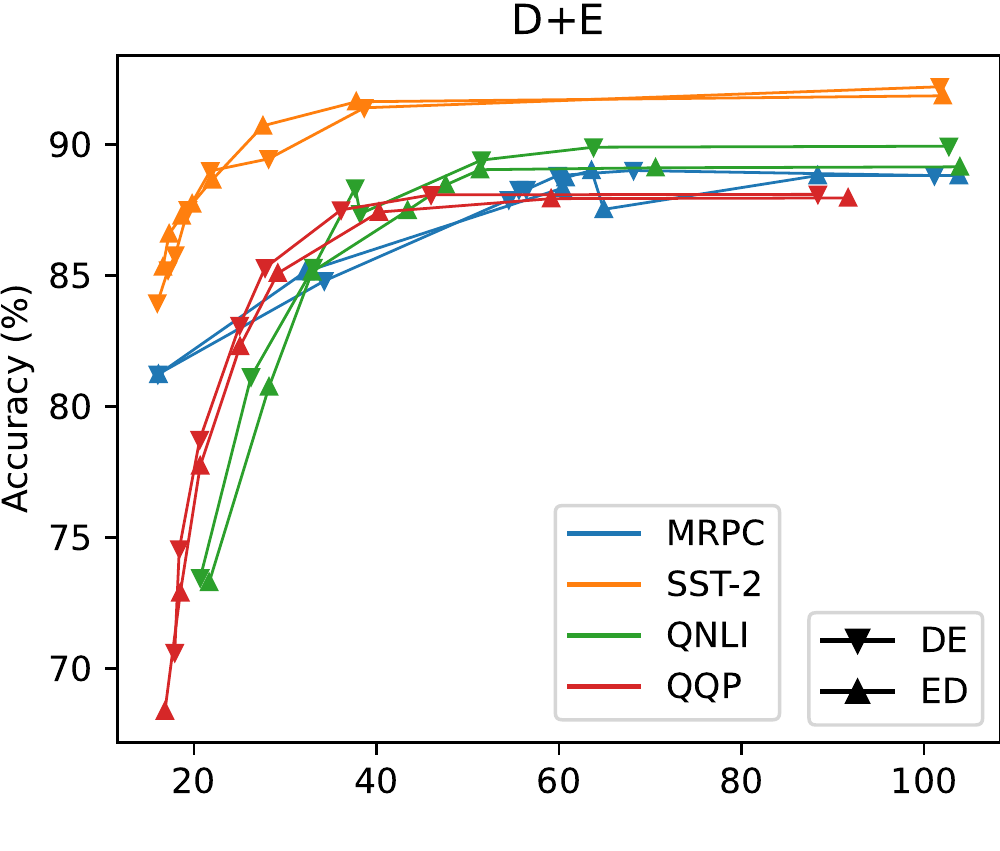}
	\includegraphics[width=0.33\textwidth]{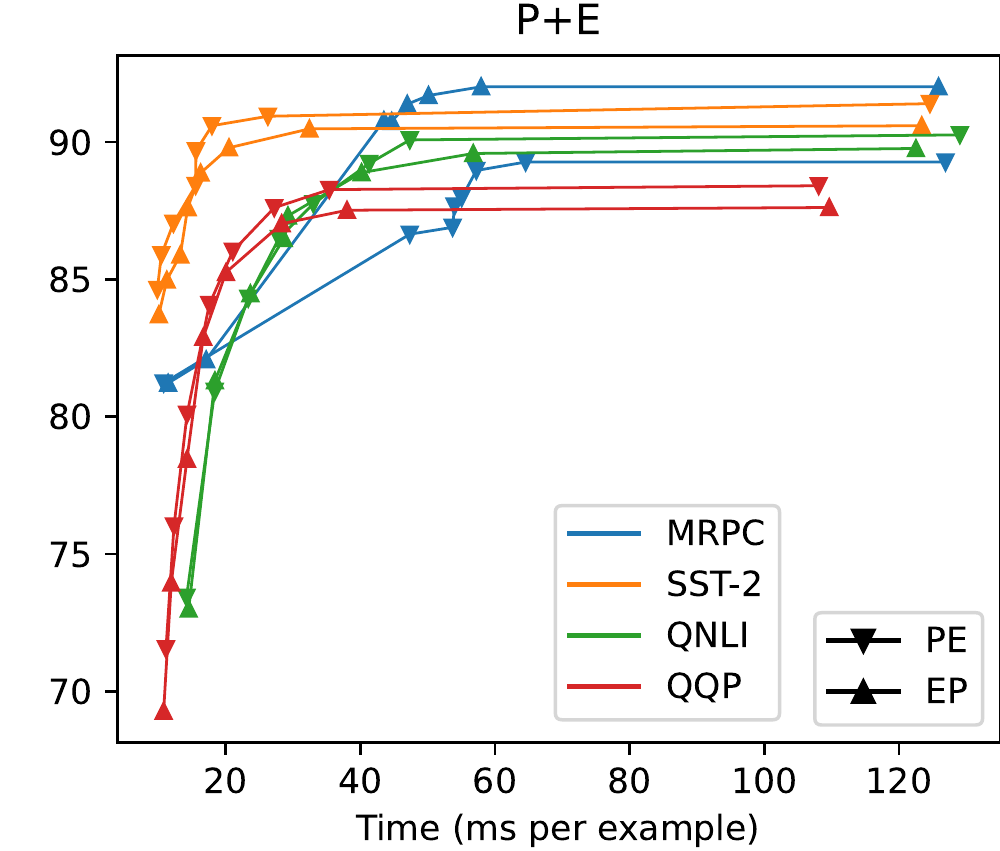}
	\includegraphics[width=0.33\textwidth]{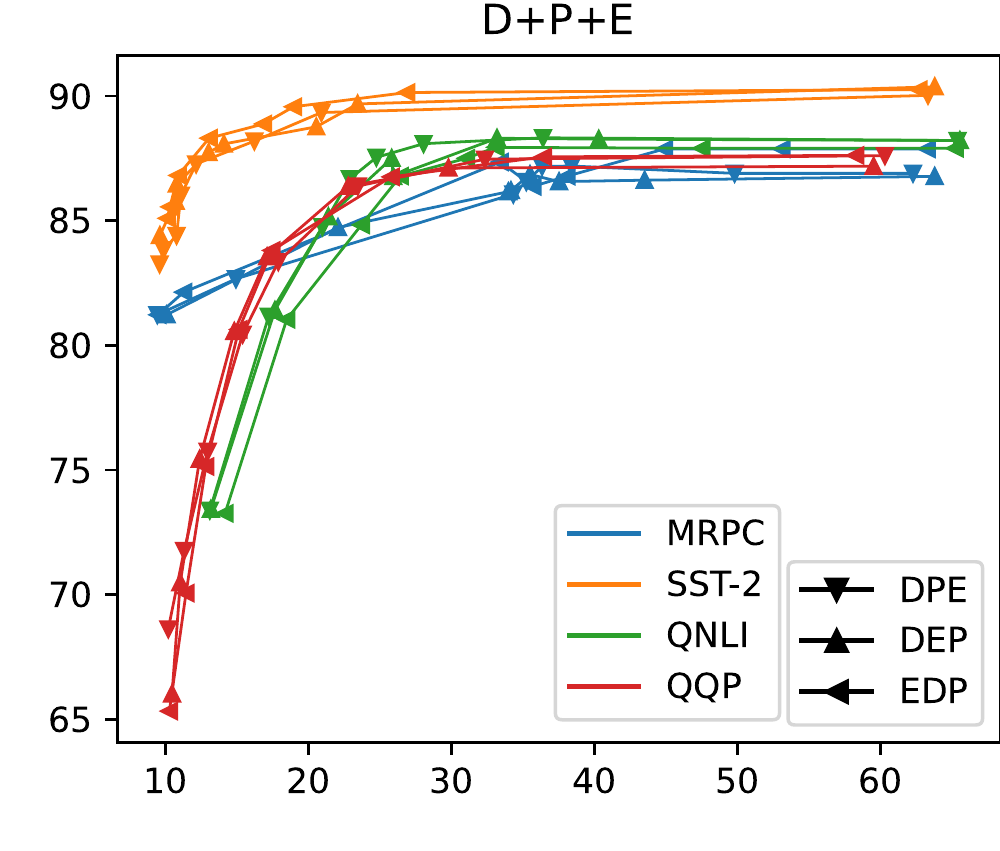}
	\caption{Different orderings of the same set of operators have similar tradeoff curves. The title of each subfigure shows the set of operators; each color represents a dataset; each marker shape represents a component ordering.}
	\label{fig:all_pipelines}
\end{figure*}

\section{Experimental Setup}
\label{sec:setup}

We introduce in this section the setups for our experiments and notations to facilitate discussions.

\subsection{Datasets and Implementation}

We conduct experiments with the RoBERTa-base model ~\cite{roberta} on four sequence classification tasks:\ MRPC~\cite{mrpc}, SST-2~\cite{sst}, QNLI~\cite{squad,glue}, and QQP~\cite{qqp}.
Our implementation of efficiency methods are adopted from Transformers~\cite{huggingface}, Fastformers~\cite{fastformers}, and DeeBERT~\cite{deebert}.
We train all the models with an NVIDIA Tesla T4 GPU.
We evaluate them with an AMD Ryzen 5800X CPU, where we measure the wall-clock time for inference.

\subsection{Pipelines and Operators}

In all experiments, we represent a pipeline with a string of bald capital letters, where each letter represents an efficiency operator, and the order of these letters represents their order.

The operators include:\
\repr{D}istillation, Structured \repr{P}runing, \repr{E}arly Exiting, Dynamic \repr{L}ength, and \repr{Q}uantization.
For example, the string ``\repr{DEPLQ}'' represents a pipeline of sequentially applying the following operators to a fine-tuned model: (1) distill it into a student model; (2) add early exiting classifiers to it and train; (3) apply structured pruning to make each layer ``thinner'' and distill from the unpruned model; and (4) use dynamic length and quantization for the final inference.
Additionally, we use \repr{O} to represent an ``empty'' pipeline, i.e., directly applying the \textbf{O}riginal fine-tuned model.

Before experimenting with pipelines, we explore the optimal setting (e.g., learning rate, batch size) for each individual operator and use the same setting in the pipelines.
This is a realistic approach since it is impractical to search for the optimal setting for every component in every new pipeline.
We use a learning rate of $10^{-5}$ and a batch size of $8$ in all training.
All training procedures, including original fine-tuning, distillation, and training with early exiting, consist of $10$ epochs, with no early stopping.
For pruning, we prune the number of attention heads from $12$ to $8$ and the intermediate dimension from $3072$ to $1536$, since in our preliminary experiments, this combination is a sweet spot on the Pareto frontier.

\begin{figure*}[ht]
	\includegraphics[width=0.33\textwidth]{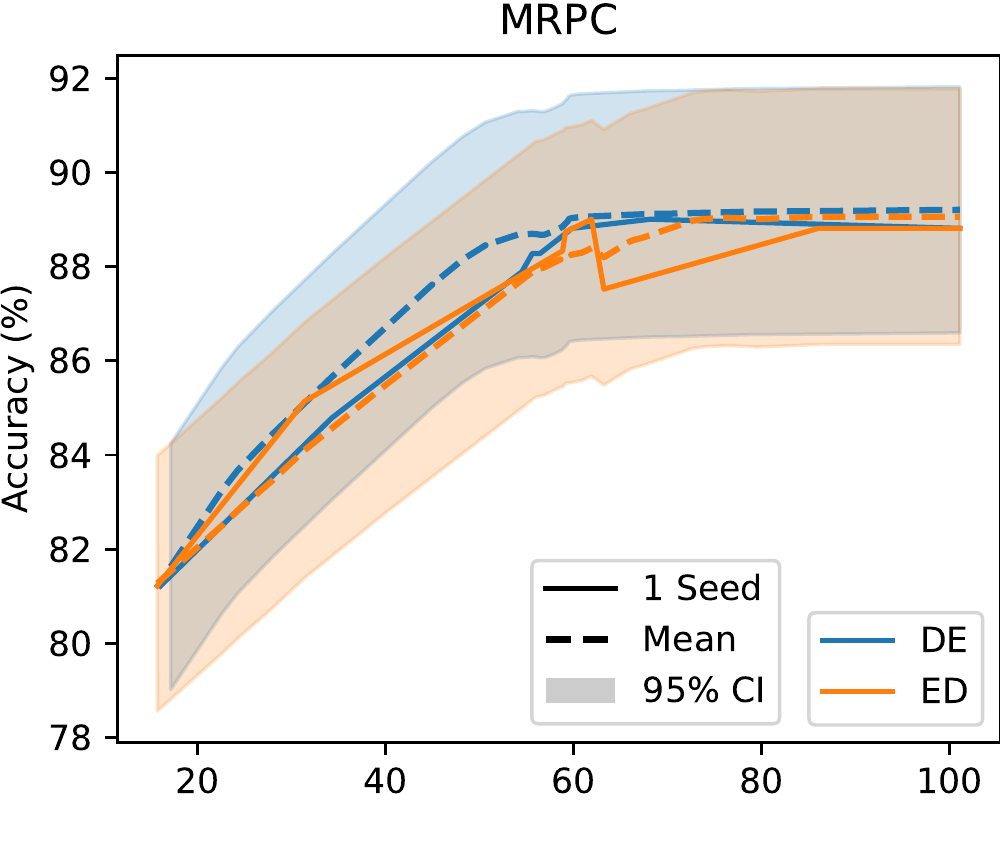}
	\includegraphics[width=0.33\textwidth]{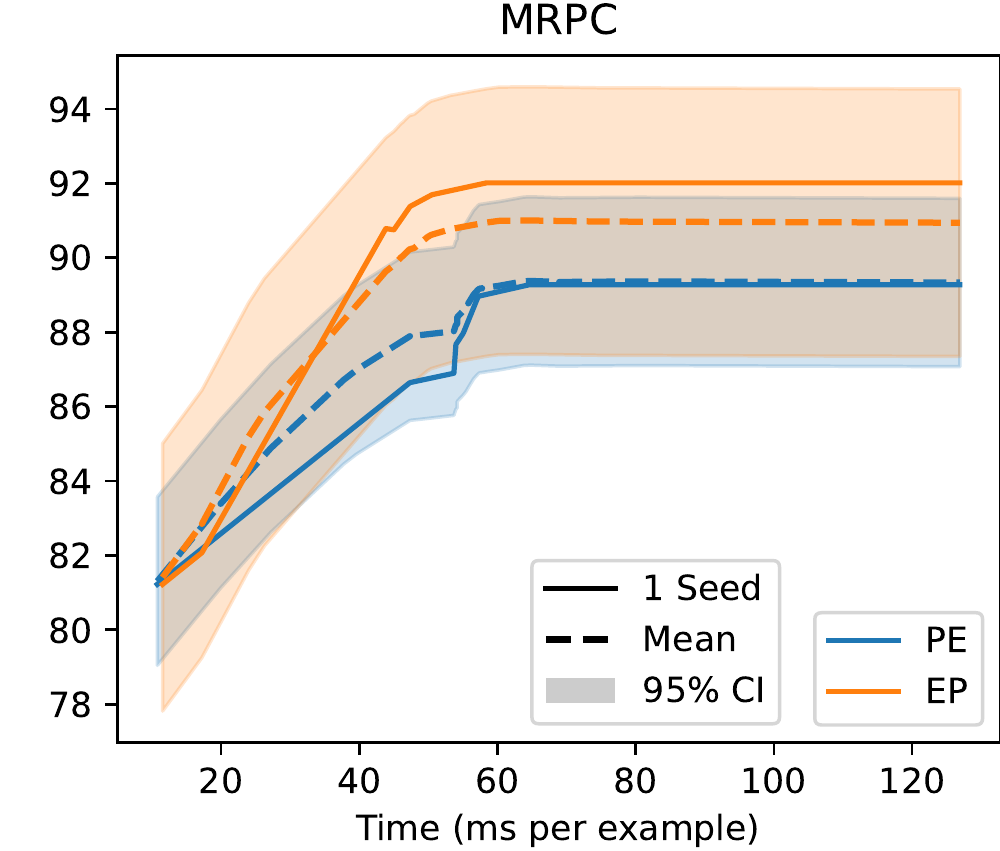}
	\includegraphics[width=0.33\textwidth]{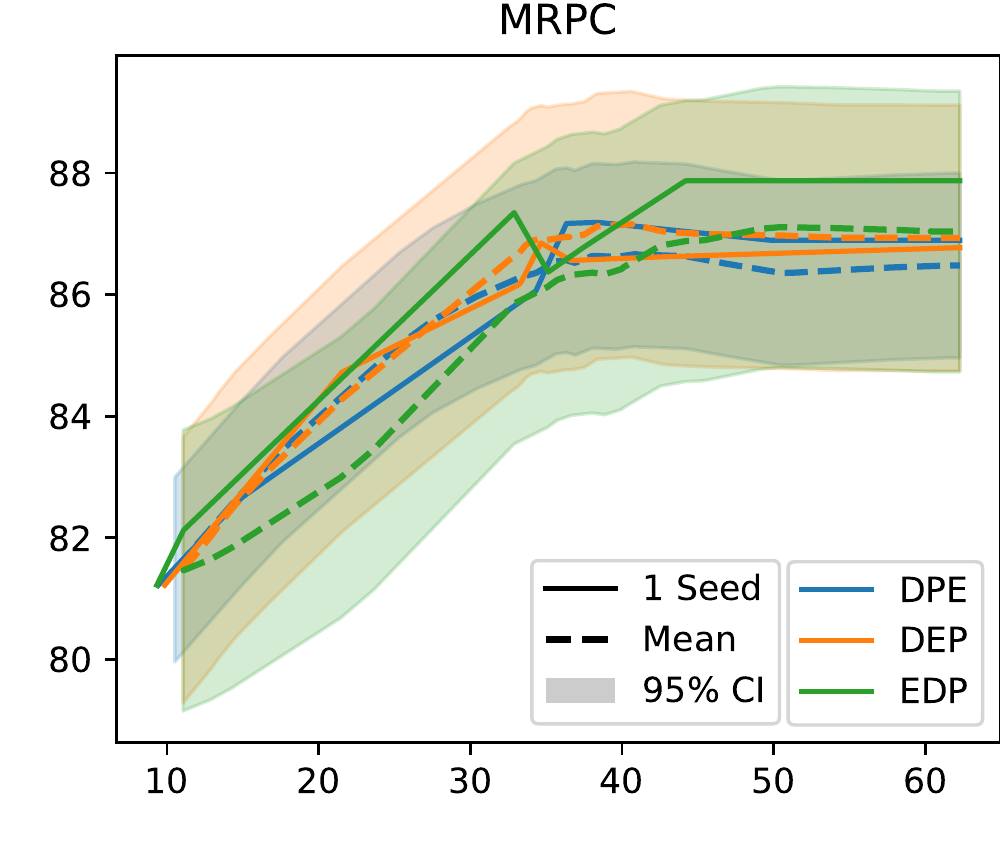}
	\caption{Comparing the results of a single run (solid lines; same as the ones from \Cref{fig:all_pipelines}) and the results from multiple runs (dashed lines for the mean and shaded areas for 95\% confidence intervals).
    }
	\label{fig:randomness}
\end{figure*}

\section{Operator Commutativity and Order}
\label{sec:exp-commutative}

Given a set of operators, we naturally wonder about the best order to apply them.
Although this question seems formidable due to the exponentially large number of possible orderings, we show that the question is actually simpler than expected.

\subsection{Limited Candidate Space}
Among the operators discussed in this paper, \repr{D}, \repr{P}, and \repr{E} require additional training steps, while \repr{Q} and \repr{L} are directly applicable right before inference.
Therefore, \repr{D}, \repr{P}, and \repr{E} (Group \rom{1}) should always appear before \repr{Q} and \repr{L} (Group \rom{2}) in the pipeline.
Moreover, applying \repr{D} after \repr{P} does not make sense, since \repr{D} initializes a small student, and the efficiency brought by the pruning step cannot be passed over to the student.
With these constraints, we greatly reduce the number of meaningful ordering candidates.

\subsection{Commutative Properties of Operators}
In this subsection, we show how commutative these operators are, i.e., how much difference swapping their orders makes.
We discuss the two groups separately.

\paragraph{Group \rom{1}}
We show the results of swapping the order of operators from Group \rom{1} in \Cref{fig:all_pipelines}.
Since early exiting is involved, which means the model can achieve different tradeoffs between accuracy and inference time, we present each ordering as a \textit{tradeoff curve}, where points are drawn by varying the early exiting threshold of confidence.
We can see that when we use the same set of operators, different orderings have similar tradeoff curves, in most cases.

Exceptions exist, however, in the \repr{E}$+$\repr{P} combination on the MRPC dataset.
We hypothesize that this is due to training randomness.
In order to reduce randomness, we repeat the experiment with additional random seeds and show in \Cref{fig:randomness} the results on MRPC, which is the smallest dataset and hence the one most influenced by randomness.
We can see that (1) the gap between the \textit{mean} curves is smaller than the gap between curves corresponding to using a single seed;
(2) the mean curve of each ordering lies within the 95\% confidence interval (95\% CI) of other orderings.
This shows that the differences between tradeoff curves of different orderings can, at least partly, be attributed to training randomness.

To further quantify the degree of dissimilarity between different orderings, we define and calculate the \textit{distance} between tradeoff curves.
The distance between two tradeoff curves is defined as the maximum accuracy ($y$-axis) difference at the same inference time ($x$-axis) point.
We compare distances between tradeoff curves (1) generated by the same operator order but with different random seeds; and (2) generated by different operator orders.
We show the results in \Cref{tab:sig}.
We can see that while tradeoff curves generated by the same operator order tend to have a smaller average distance, the difference between same/different orders is typically small and the one-standard-deviation (1-SD) intervals of both sides always overlap.
Although we are unable to find a suitable significance test since the distances are not completely independent, the above analysis shows that the difference of distances between curves from same/different orders is likely not significant.
On the other hand, in terms of \textit{absolute difference}, the difference made by swapping the order of operators is only fractionally higher than choosing a different random seed (less than $2\%$ in most cases.
Therefore in practice, we can regard these operators as commutative.

\begin{table}[!t]
\centering
\resizebox{\columnwidth}{!}{%
\begin{tabular}{llccc}
\toprule
Dataset & Order & D+E & P+E & D+P+E \\ \midrule
\multirow{2}{*}{MRPC}  & Same & $1.57\pm0.69$ & $2.31\pm0.85$ & $1.53\pm0.62$ \\
                  & Diff. & $1.74\pm0.40$ & $4.12\pm1.10$ & $2.59\pm0.97$ \\ \midrule
\multirow{2}{*}{SST-2} & Same & $1.30\pm0.39$ & $1.64\pm0.46$ & $1.49\pm0.53$ \\
                  & Diff. & $1.48\pm0.46$ & $1.84\pm0.62$ & $1.98\pm0.82$ \\ \midrule
\multirow{2}{*}{QNLI}  & Same & $2.24\pm1.20$ & $4.40\pm2.49$ & $3.41\pm2.51$ \\
                  & Diff. & $3.93\pm0.82$ & $4.58\pm2.39$ & $4.91\pm2.44$ \\ \midrule
\multirow{2}{*}{QQP}   & Same & $2.38\pm1.36$ & $2.11\pm0.86$ & $2.30\pm1.05$ \\
                  & Diff. & $3.64\pm1.14$ & $3.30\pm1.27$ & $4.56\pm1.71$ \\
 \bottomrule
\end{tabular}%
}
\caption{The mean and the standard deviation (SD) of \textit{distances} between tradeoff curves belonging to same/different orders (the same ordering is run with multiple random seeds). For all entries, the 1-SD intervals of same/different orders overlap.
}
\label{tab:sig}
\end{table}

\paragraph{Group \rom{2}}
The two operators, \repr{Q} and \repr{L}, are independent of each other and therefore their order can be arbitrarily swapped (i.e., they are strictly commutative by definition).
We show the results of applying \repr{Q} and/or \repr{L} at the end of different pipelines in \Cref{tab:ql}.
We do not report the accuracy of +\repr{L} since using dynamic length does not change the model's accuracy.

\smallskip
\noindent
Based on the above discussion, when we have a set of components to apply, it suffices to simply pick a reasonable order from the candidate space.

\begin{table*}[ht]
	\centering
	\begin{tabular}{c@{\hspace{10pt}}c@{\hspace{30pt}}cc@{\hspace{10pt}}rcccc}
		\toprule
		Dataset & Pipeline & \multicolumn{2}{|l|}{\hspace{25pt}Accuracy (\%)} & \multicolumn{5}{c}{\hspace{0pt}Time (ms per example)} \\ \midrule
		&  & \multicolumn{1}{|c}{Raw} & +\repr{Q} (relative diff.) & \multicolumn{1}{|c}{Raw} & +\repr{Q} & +\repr{L} & +\repr{QL} & \multicolumn{1}{c}{+\repr{QL} (est.)}\\ \midrule
\multirow{4}{*}{MRPC}  & \repr{O} & 92.7 & 92.5 ($-$0.2\%) & 170.7 & $-$50\% & $-$83\% & $-$94\% & $-$92\% \\
		& \repr{D}                & 89.2 & 88.8 ($-$0.4\%) &  85.5 & $-$49\% & $-$82\% & $-$94\% & $-$91\%  \\
		& \repr{P}                & 91.0 & 89.0 ($-$2.2\%) & 122.4 & $-$64\% & $-$86\% & $-$94\% & $-$95\%  \\
		& \repr{DP}               & 88.9 & 87.9 ($-$1.1\%) &  59.3 & $-$62\% & $-$84\% & $-$94\% & $-$94\%  \\ \midrule
\multirow{4}{*}{SST-2} & \repr{O} & 93.7 & 93.5 ($-0.2$\%) & 170.8 & $-$50\% & $-$86\% & $-$97\% & $-$93\%  \\
		& \repr{D}                & 92.3 & 92.3 ($-$0.0\%) &  85.5 & $-$49\% & $-$86\% & $-$97\% & $-$93\%  \\
		& \repr{P}                & 92.4 & 91.7 ($-$0.8\%) & 126.7 & $-$66\% & $-$89\% & $-$97\% & $-$96\%  \\
		& \repr{DP}               & 92.0 & 90.9 ($-$1.2\%) &  62.9 & $-$65\% & $-$88\% & $-$97\% & $-$96\%  \\ \midrule
\multirow{4}{*}{QNLI}  & \repr{O} & 92.3 & 92.1 ($-$0.2\%) & 174.2 & $-$51\% & $-$83\% & $-$95\% & $-$92\%  \\
		& \repr{D}                & 91.3 & 90.7 ($-$0.7\%) &  86.9 & $-$50\% & $-$82\% & $-$95\% & $-$91\%  \\
		& \repr{P}                & 91.5 & 91.4 ($-$0.1\%) & 121.5 & $-$64\% & $-$86\% & $-$95\% & $-$95\%  \\
		& \repr{DP}               & 89.8 & 89.6 ($-$0.2\%) &  62.6 & $-$65\% & $-$85\% & $-$95\% & $-$95\%  \\ \midrule
\multirow{4}{*}{QQP}   & \repr{O} & 88.6 & 88.3 ($-$0.3\%) & 172.3 & $-$51\% & $-$86\% & $-$96\% & $-$93\%  \\
		& \repr{D}                & 87.9 & 87.7 ($-$0.2\%) &  88.2 & $-$51\% & $-$85\% & $-$97\% & $-$93\%  \\
		& \repr{P}                & 88.5 & 88.5 ($-$0.0\%) & 118.3 & $-$63\% & $-$87\% & $-$97\% & $-$95\%  \\
		& \repr{DP}               & 87.6 & 87.6 ($-$0.0\%) &  58.8 & $-$62\% & $-$86\% & $-$97\% & $-$95\%  \\ \bottomrule
	\end{tabular}%
	\caption{Accuracy drops and time savings provided by quantization (\repr{Q}) and dynamic length inference (\repr{L}) applied at the end of pipelines. The accuracy drops and time savings of most operators are cumulative.}
	\label{tab:ql}
\end{table*}

\begin{figure*}[ht]
	\includegraphics[width=0.33\textwidth]{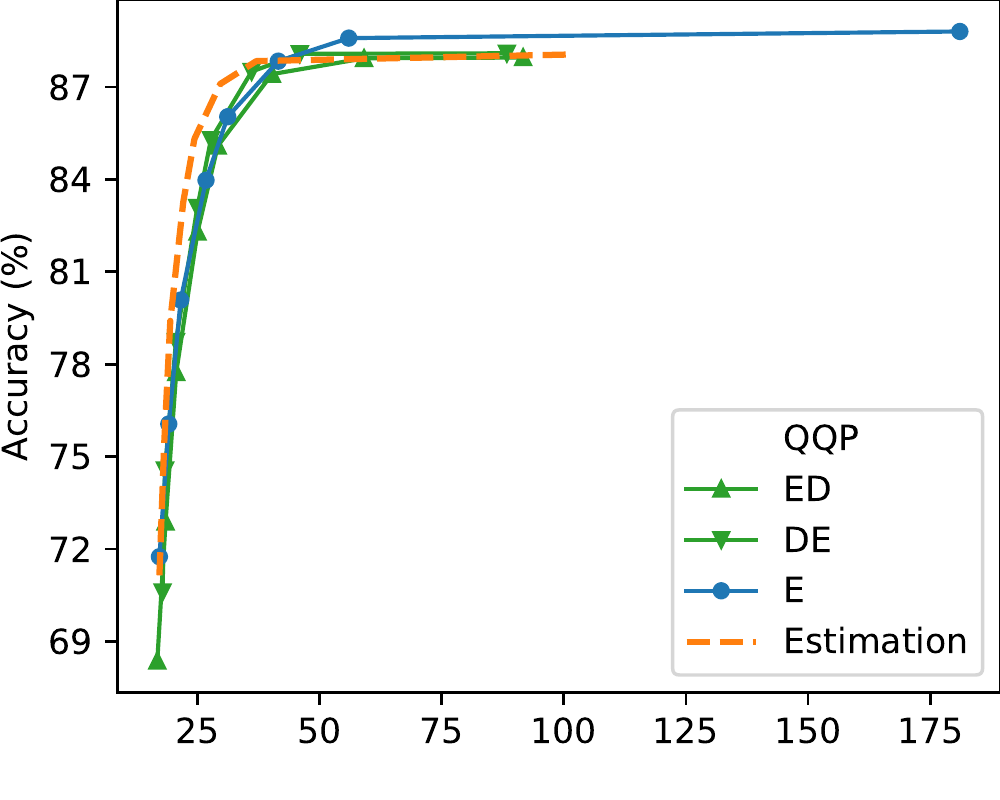}
	\includegraphics[width=0.33\textwidth]{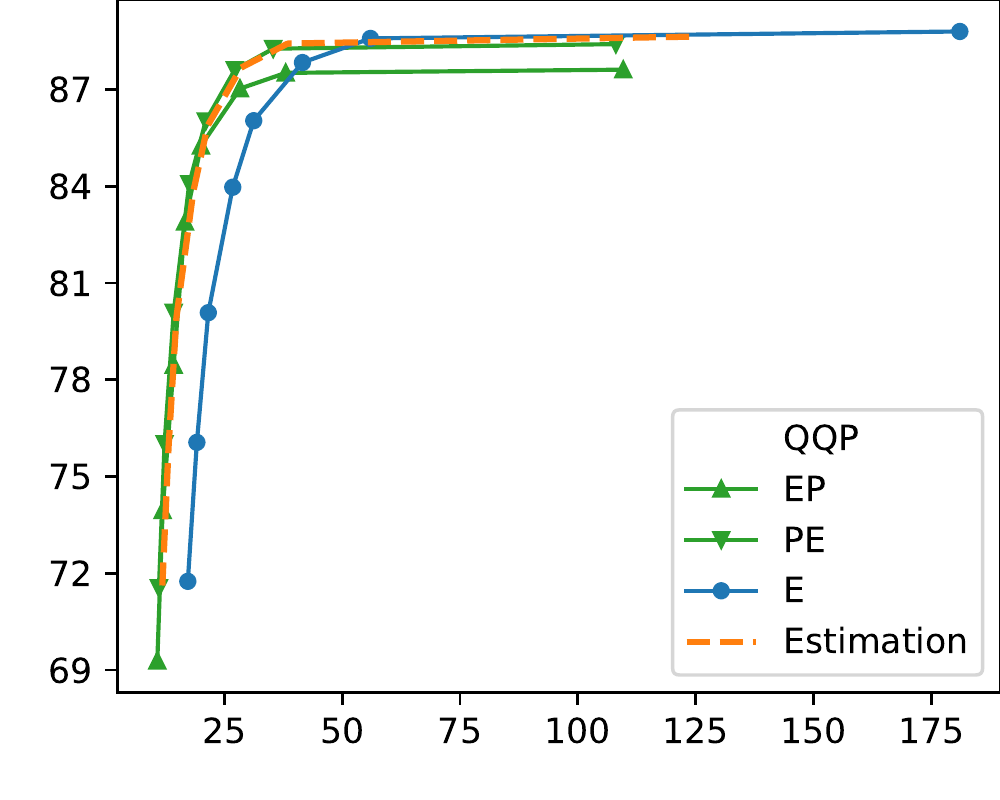}
	\includegraphics[width=0.33\textwidth]{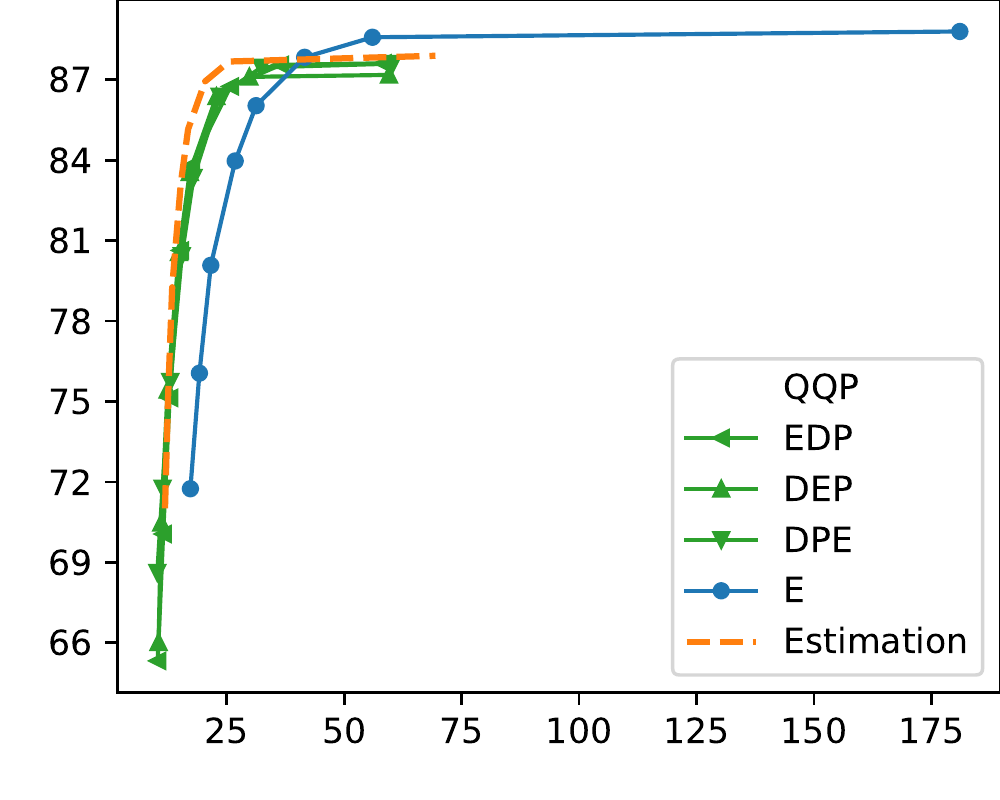}
	\includegraphics[width=0.33\textwidth]{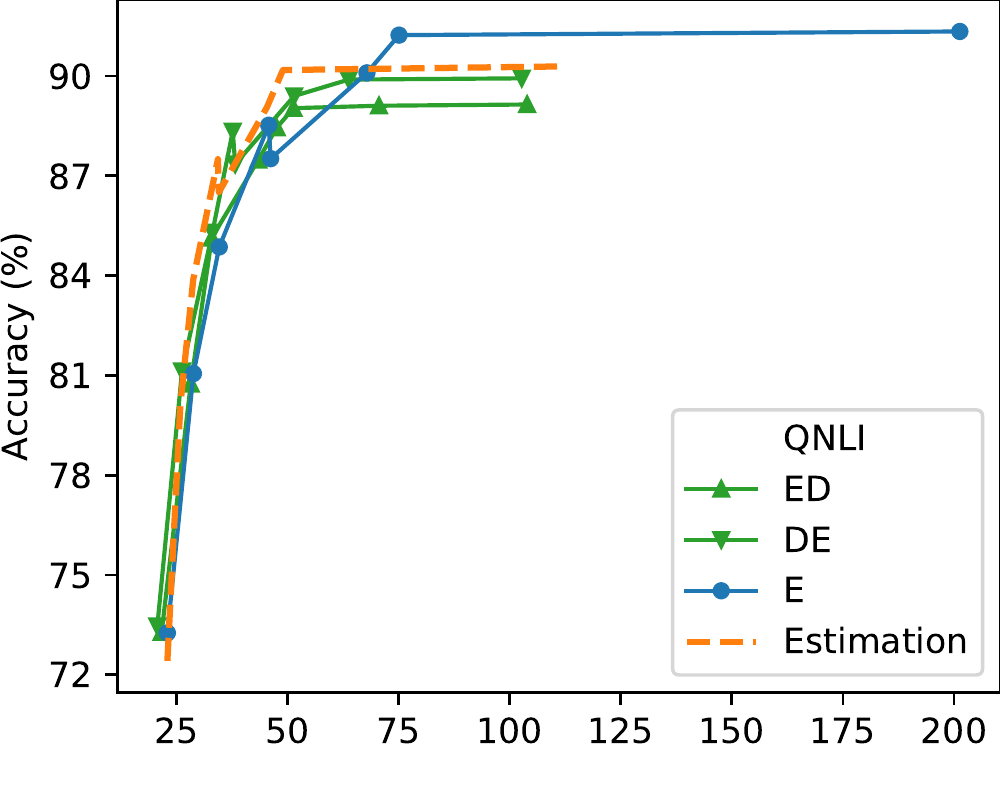}
	\includegraphics[width=0.33\textwidth]{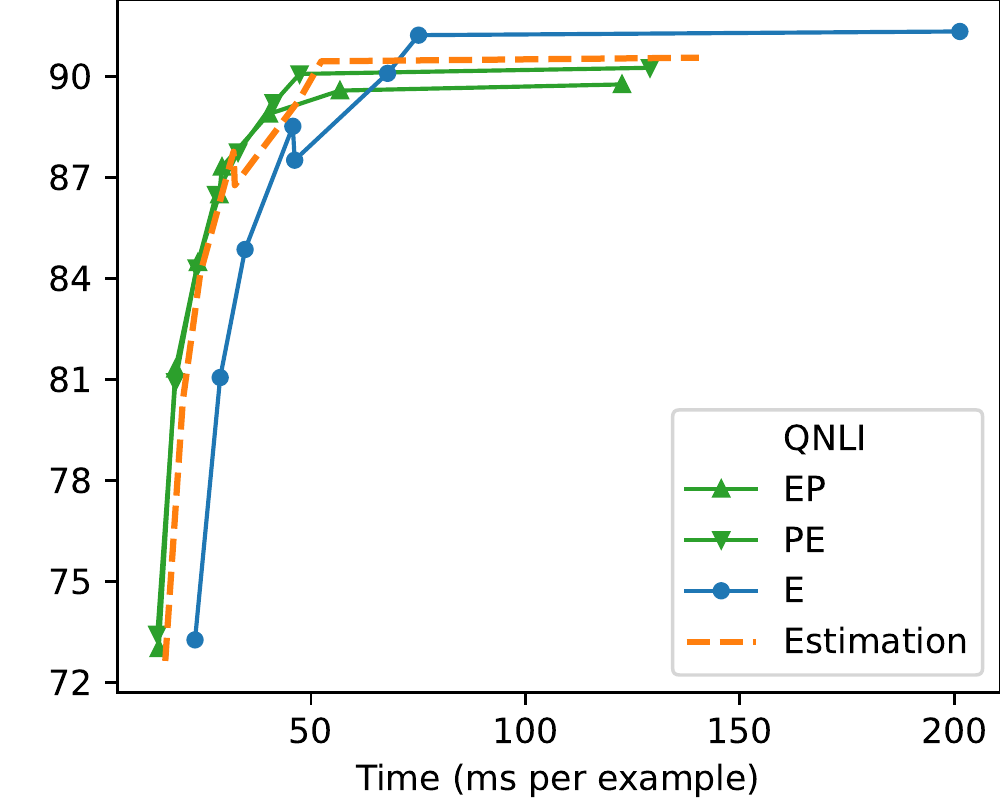}
	\includegraphics[width=0.33\textwidth]{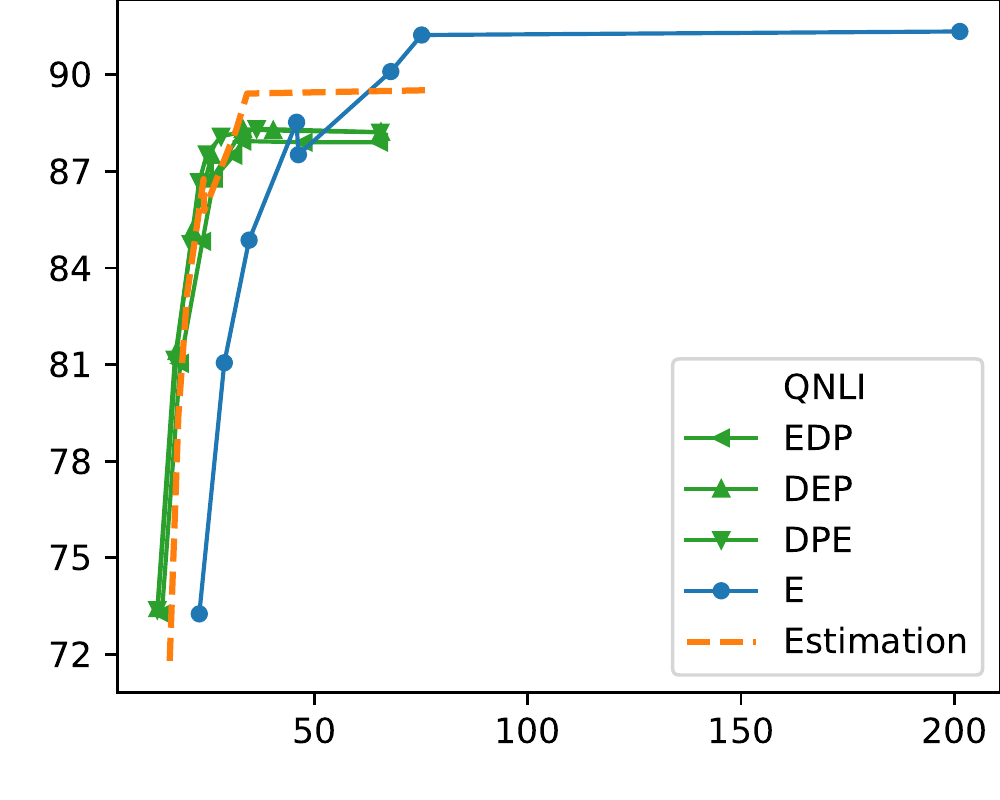}
	\caption{Estimating the tradeoff curves of target pipelines based on the results of individually applying operators. Green curves:\ measured tradeoff curves of target pipelines; blue curves:\ measured tradeoff curves of individually applying the operator \repr{E}; orange curves:\ estimated tradeoff curves for the target pipelines.}
	\label{fig:estimate}
\end{figure*}

\section{Operator Cumulativeness and Predictability of Pipelines}

In order to choose components for an efficiency pipeline, an important question is whether time savings and accuracy drops of individual operators are cumulative.
In this subsection, we show that they are indeed cumulative to the degree that accuracy--efficiency tradeoffs of a new pipeline can be estimated by combining the results of individual operators.

We first discuss operators from Group \rom{1}.
In \Cref{fig:estimate}, we show how we can estimate the tradeoff curve of a new pipeline
based on the results of its constituents, using the two larger and more stable datasets, QQP and QNLI.
For example, in the top-right subfigure, we show the estimation for the tradeoff curves of pipelines comprising \repr{E}, \repr{D}, and \repr{P}, based on the results of individually applying each of these operators.

The idea for estimating accuracy drops is based on the following cumulativeness assumption.
Suppose \repr{R} is a pipeline, $A_\repr{*}$ is the accuracy for a pipeline \repr{*},
then we assume
\begin{eqnarray}
	A_{\repr{R+D}} &=&
	\frac{A_{\repr{D}}}{A_{\repr{O}}} \times A_{\repr{R}}, \\
	A_{\repr{R+P}} &=&
	\frac{A_{\repr{P}}}{A_{\repr{O}}} \times A_{\repr{R}}.
	\label{eqn:adjustment}
\end{eqnarray}
In other words, our assumption is that adding \repr{D} or \repr{P} to \textit{any} pipeline should result in similar relative accuracy drops.
We can therefore estimate the accuracy of \repr{ED}, \repr{EP}, and \repr{EDP} (and other orders of the same set of operators) as follows:
(1) calculate accuracy drops of \repr{D} and \repr{P} relative to \repr{O};
(2) multiply the relative accuracy drops to points on \repr{E}'s tradeoff curve.

The idea for estimating time savings is also similar, but additional modifications are necessary:
\begin{itemize}

\item When we add \repr{P} to \repr{E}, since they work on reducing different dimensions of the model (width and depth), the time savings are independent and directly cumulative:
\begin{equation}
	T_{\repr{E}+\repr{P}} =
	\frac{T_{\repr{P}}}{T_{\repr{O}}} \times T_{\repr{E}},
\end{equation}
where similarly, $T_\repr{*}$ is the inference time for a pipeline \repr{*}.

\item When we add \repr{D} to \repr{E}, we need to consider the fact that both \repr{D} and \repr{E} reduce the number of layers.
When the early exiting threshold is extremely large and the model uses all layers for inference, the relative time saving will be close to $T_{\repr{D}}/T_{\repr{O}}$;
when the early exiting threshold is extremely small and the model exits after only one layer, adding \repr{D} provides no extra time saving.
Therefore, for non-extreme cases, we can estimate the time saving for \repr{E+D} by interpolating the above two extreme cases:
\begin{equation}
	T_{\repr{E}+\repr{D}} = t_\repr{E} + (T_\repr{E} - t_\repr{E}) \times \frac{T_{\repr{D}}}{T_{\repr{O}}},
\end{equation}
where $t_\repr{E}$ is the minimum value of time in the tradeoff curve of \repr{E} (i.e., the point where we early exit after only one layer).

\item When we add both \repr{P} and \repr{D} to \repr{E}, we combine the above two estimations:
\begin{equation}
    T_{\repr{E}+\repr{DP}} = \Big(t_\repr{E} + (T_\repr{E} - t_\repr{E}) \times \frac{T_{\repr{D}}}{T_{\repr{O}}} \Big)
    \times \frac{T_{\repr{P}}}{T_{\repr{O}}}.
\end{equation}

\end{itemize}

\noindent
We use the above ideas to estimate tradeoff curves of new pipelines and show the results in \Cref{fig:estimate}.
From the figure, we can see that the estimation curves (orange) align well with the measured curves (green), across different datasets and operator sets, showing that individual components from Group \rom{1} are cumulative with each other under these settings.

For operators from Group \rom{2}, we refer to \Cref{tab:ql}.
We see that on the same dataset, \repr{Q} leads to similar accuracy drops when added to any pipeline, especially on the larger and more stable datasets, QNLI and QQP.
Time savings, on the other hand, are trickier:
\begin{itemize}

\item \repr{L} provides consistent time savings for all pipelines, showing that it is cumulative with any operator from Group \rom{1}.

\item \repr{L} and \repr{Q} are also cumulative with each other, as evidenced by the fact that the measured time savings of +\repr{QL} align well with the estimation of +\repr{QL}, which is simply multiplying the respective savings of \repr{Q} and \repr{L}.

\item \repr{Q}, however, is cumulative only with \repr{D} and \repr{E}, but not \repr{P}---it saves more time for pipelines with \repr{P}.
This is because quantization's acceleration is different for different types of operations, and pruning changes the proportion of each type of operations within a transformer layer, while distillation or early exiting does not.
When we estimate the tradeoff of a pipeline containing both \repr{Q} and \repr{P}, \repr{PQ} needs to be treated as a compound operator, and it is cumulative with others.
This also applies to other operators that change the connection within a transformer layer.

\end{itemize}

\noindent
The observation that operators are cumulative facilitates future experiments on efficiency pipelines.
For pipelines that are tedious to train and evaluate, simply measuring the performance of their components can provide us with a reliable estimation of the pipeline's behavior.

\section{Conclusion}

In this paper, we consider efficiency methods as operators applied on transformer models and study the properties of these operators.
We observe from experiments that (1) operators are commutative:\ changing their order has little practical impact on the final efficiency--accuracy tradeoff; (2) operators are cumulative:\ a new pipeline's performance can be estimated by cumulating time savings and accuracy drops of each component.
These observations facilitate future construction of efficiency pipelines and also provide an interesting direction to better understand efficiency pipelines.

\section*{Acknowledgements}
This research is supported in part by the Canada First Research Excellence Fund and the Natural Sciences and Engineering Research Council (NSERC) of Canada.

\bibliography{tacl}
\bibliographystyle{acl_natbib}


\end{document}